\documentclass[11pt,a4paper]{article}
\usepackage{hvfloat}

\usepackage{rotating,amsmath,booktabs,graphicx}
\usepackage{lscape}
\usepackage[table]{xcolor}
\usepackage[hyperref]{acl2021}
\usepackage{times}
\usepackage{latexsym}
\usepackage{amsmath,graphicx}
\usepackage{bm}
\usepackage{multirow}
\usepackage{url}
\usepackage{rotating}
\usepackage{pdflscape}
\usepackage{comment}
\usepackage{amssymb}
\usepackage{stfloats}
\usepackage{tabu}
\usepackage{bbm}
\usepackage{array}
\usepackage{hyperref}
\usepackage{footnote}
\usepackage{booktabs}
\usepackage[flushleft]{threeparttable}
\makesavenoteenv{tabular}
\makesavenoteenv{table}
\usepackage{tablefootnote}
\usepackage{graphicx}
\usepackage{caption}
\usepackage{subcaption}
\usepackage{tikz}
\usetikzlibrary{patterns}
\usetikzlibrary{positioning}

\tikzset{
    slope/.store in=\slope,
    pattern color/.append code={\def\patterncolor{#1}}
}

\newcommand{\theslope}{0.7}

\pgfdeclarepatternformonly{diagonal lines}
{\pgfpoint{-.1mm/\theslope}{-.1mm}}{\pgfpoint{1.1mm/\theslope}{1.1mm}}
{\pgfpoint{1mm/\theslope}{1mm}}
{
    \pgfsetlinewidth{0.4pt}
    \pgfpathmoveto{\pgfpoint{-.1mm/\theslope}{-.1mm}}
    \pgfpathlineto{\pgfpoint{1.1mm/\theslope}{1.1mm}}
    \pgfusepath{stroke}
}

\pgfdeclarepatternformonly[\slope,\patterncolor]{slant lines}
{\pgfpoint{-.1mm/\slope}{-.1mm}}{\pgfpoint{1.1mm/\slope}{1.1mm}}
{\pgfpoint{1mm/\slope}{1mm}}
{
    \pgfsetlinewidth{0.4pt}
    \pgfpathmoveto{\pgfpoint{-.1mm/\slope}{-.1mm}}
    \pgfpathlineto{\pgfpoint{1.1mm/\slope}{1.1mm}}
    \pgfsetstrokecolor{\patterncolor}
    \pgfusepath{stroke}
}

\newcommand{\PreserveBackslash}[1]{\let\temp=\\#1\let\\=\temp}
\newcolumntype{C}[1]{>{\PreserveBackslash\centering}p{#1}}
\newcolumntype{R}[1]{>{\PreserveBackslash\raggedleft}p{#1}}
\newcolumntype{L}[1]{>{\PreserveBackslash\raggedright}p{#1}}

\newcommand\blfootnote[1]{%
  \begingroup
  \renewcommand\thefootnote{}\footnote{#1}%
  \addtocounter{footnote}{-1}%
  \endgroup
}

\newcommand{\pgftextcircled}[1]{
    \setbox0=\hbox{#1}%
    \dimen0\wd0%
    \divide\dimen0 by 2%
    \begin{tikzpicture}[baseline=(a.base)]%
        \useasboundingbox (-\the\dimen0,0pt) rectangle (\the\dimen0,1pt);
        \node[circle,draw,outer sep=0pt,inner sep=0.1ex] (a) {#1};
    \end{tikzpicture}
}
\let\textcircled=\pgftextcircled


\graphicspath{ {images/} }
\aclfinalcopy 

\definecolor{gainsboro}{rgb}{0.86, 0.86, 0.86}
\newcommand{\wx}[1]{\textcolor{magenta}{\bf\small [#1 --WX]}}

\newcommand{\doubleRowCell}[3]{
  \begin{tabular}{@{}c@{}}
                  \fontsize{10}{13}\selectfont  #1\\ [-0.4em]
                  \fontsize{7}{2}\selectfont #2 / #3\\
                 \end{tabular}
}

\title{Neural semi-Markov CRF for Monolingual Word Alignment}

\author{Wuwei Lan$^\bigstar$\textsuperscript{1}, Chao Jiang$^\bigstar$\textsuperscript{2}, Wei Xu\textsuperscript{2} \\
\textsuperscript{1} Department of Computer Science and Engineering, Ohio State University\\
\textsuperscript{2} School of Interactive Computing, Georgia Institute of Technology\\
  {\small \tt lan.105@osu.edu   \{chao.jiang, wei.xu\}@cc.gatech.edu}
}

\begin{document}
\maketitle
\begin{abstract}

Monolingual word alignment is important for studying fine-grained editing operations (i.e., deletion, addition, and substitution) in text-to-text generation tasks, such as paraphrase generation, text simplification, neutralizing biased language, etc.  In this paper, we present a novel  neural semi-Markov CRF alignment model, which unifies word and phrase alignments through variable-length spans. We also create a new benchmark with human annotations that cover four different text genres to evaluate monolingual word alignment models in more realistic settings.  Experimental results show that our proposed model outperforms all previous approaches for monolingual word alignment as well as a competitive QA-based baseline, which was previously only applied to bilingual data. Our model demonstrates good generalizability to three out-of-domain datasets and shows great utility in two downstream applications: automatic text simplifi-cation and sentence pair classification tasks.\footnote{Our code and data will be available at: \url{https://github.com/chaojiang06/neural-Jacana}}
\end{abstract}
\vspace{-1cm}

\blfootnote{$^\bigstar$Authors contributed equally.}

\section{Introduction}

\begin{figure}
    \centering
    \fbox{\includegraphics[width=0.96\linewidth]{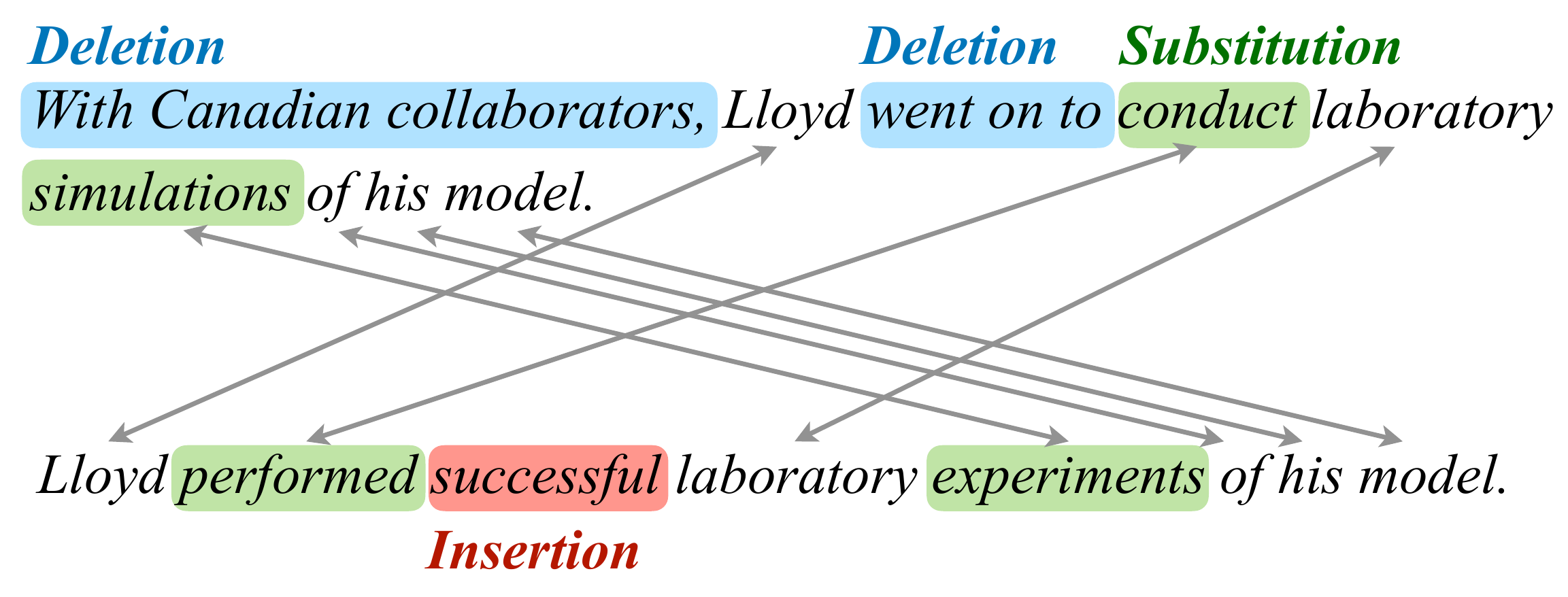}}
    \caption{An example that illustrates monolingual word alignment (shown as arrows) can support analysis of human editing process and training of text generation models (\S \ref{sec:text_simplification}), such as for simplifying complex sentences for children to read.}
    \label{fig:word_alignment_samples}
    \vspace{-0.5cm}
\end{figure}

Monolingual word alignment aims to align words or phrases with similar meaning in two sentences that are written in the same language. It is useful for improving the interpretability in natural language understanding tasks, including semantic textual similarity \cite{li-srikumar-2016-exploiting} and question answering \cite{yao-phd-2014}. Monolingual word alignment can also support the analysis of human editing operations (Figure \ref{fig:word_alignment_samples}) and improve model performance for text-to-text generation tasks, such as text simplification \cite{maddela2020controllable} and neutralizing biased language \cite{pryzant2020automatically}. It has also been shown to be helpful for data augmentation and label projection \cite{10.1162/tacl_a_00380} when combined with paraphrase generation.

One major challenge for automatic alignment is the need to handle not only alignments between words and linguistic phrases (e.g., \textit{a dozen} $\leftrightarrow$ \textit{more than 10}), but also non-linguistic phrases that are semantically related given the context (e.g., \textit{tensions} $\leftrightarrow$ \textit{relations being strained} in Figure \ref{fig:example}). In this paper, we present a novel neural semi-Markov CRF alignment model, which unifies both word and phrase alignments though variable-length spans, calculates span-based semantic similarities, and takes alignment label transitions into consideration. We also create a new manually annotated benchmark, \textbf{Multi}-Genre \textbf{M}onolingual \textbf{W}ord \textbf{A}lignment (MultiMWA), which consists of four datasets across different text genres and is large enough to support the training of neural-based models (Table \ref{statistics_on_datasets}). It addresses the shortcomings of existing datasets for monolingual word alignment: MTReference \cite{yao-phd-2014} was annotated by crowd-sourcing workers and contains many obvious errors (more details in \S \ref{sec:benchmark}); iSTS \cite{agirre2016semeval_task2} and SPADE/ESPADA \cite{arase-tsujii-2018-spade, arase-tsujii-2020-compositional} were annotated based on chunking and parsing results, which may restrict the granularity and flexibility of the alignments.  

Our experimental results show that the proposed semi-Markov CRF model achieves state-of-the-art performance with higher precision, in comparison to the previous monolingual word alignment models \cite{yao-jacana-wordalign-acl2013, yao-jacana-phrase-align2013, sultan2014back}, as well as another very competitive span-based neural model \cite{nagata-etal-2020-supervised} that had previously only applied to bilingual data. Our model exceeds 90\% F1 in the in-domain evaluation and also has very good generalizability on three out-of-domain datasets. We present a detailed ablation and error analysis to better understand the performance gains. Finally, we demonstrate the utility of monolingual word alignment in two downstream applications, namely automatic text simplification and sentence pair classification.

\section{Related Work}
Word alignment has a long history and was first proposed for statistical machine translation. The most representative ones are the IBM models\cite{Brown:1993}, which are a sequence of unsupervised models with increased complexity and implemented the GIZA++ toolkit \cite{och03:asc}. Many more works followed, such as FastAlign \cite{dyer-etal-2013-simple}. \citet{dyer-etal-2011-unsupervised} also used a globally normalized log-linear model for discriminative word alignment. \citet{bansal-etal-2011-gappy} proposed a hidden semi-Markov model to handle both continuous and noncontinuous phrase alignment. These statistical methods promoted the development of monolingual word alignment \cite{maccartney2008phrase, thadani-mckeown-2011-optimal, thadani2012joint}. \citet{yao-jacana-wordalign-acl2013} proposed a CRF aligner following \cite{blunsom2006discriminative}, then extended it to a semi-CRF model for phrase-level alignments \cite{yao-jacana-phrase-align2013}. \citet{sultan2014back} designed a simple system with heuristic rules based on word similarity and contextual evidence. 

Neural methods have been explored in the past decade primarily for bilingual word alignment. Some early attempts \cite{yang2013word,tamura2014recurrent} did not match the performance of GIZA++, but recent Transformer-based models started to outperform. \citet{garg-etal-2019-jointly} proposed a multi-task framework for machine translation and word alignment, while \citet{zenkel2020end} designed an alignment layer on top of Transformer for machine translation. Both can be trained without word alignment annotations but rely on millions of bilingual sentence pairs. As for supervised methods, \citet{stengel-eskin-etal-2019-discriminative} extracted representations from the Transformer-based MT system, then used convolutional neural network to incorporate neighboring words for alignment. \citet{nagata-etal-2020-supervised} proposed a span prediction method and formulated bilingual word alignment as a SQuAD-style question answering task, then solved it by fine-tuning multilingual BERT. We adapt their method to monolingual word alignment as a new state-of-the-art baseline (\S \ref{sec:baseline}). Some monolingual neural models have different settings from this work. \citet{ouyang-mckeown-2019-neural} introduced pointer networks for long, sentence- or clause-level alignments. \citet{arase-tsujii-2017-monolingual,arase-tsujii-2020-compositional} utilized constituency parsers for compositional and non-compositional phrase alignments. \citet{10.1162/tacl_a_00380} considered span alignment for FrameNet \cite{baker1998berkeley} annotations and treated each span pair as independent prediction.

\section{Neural Semi-CRF Alignment Model}
\label{sec:neuralcrf}

\begin{figure*}
    \centering
    \includegraphics[width=0.93\textwidth]{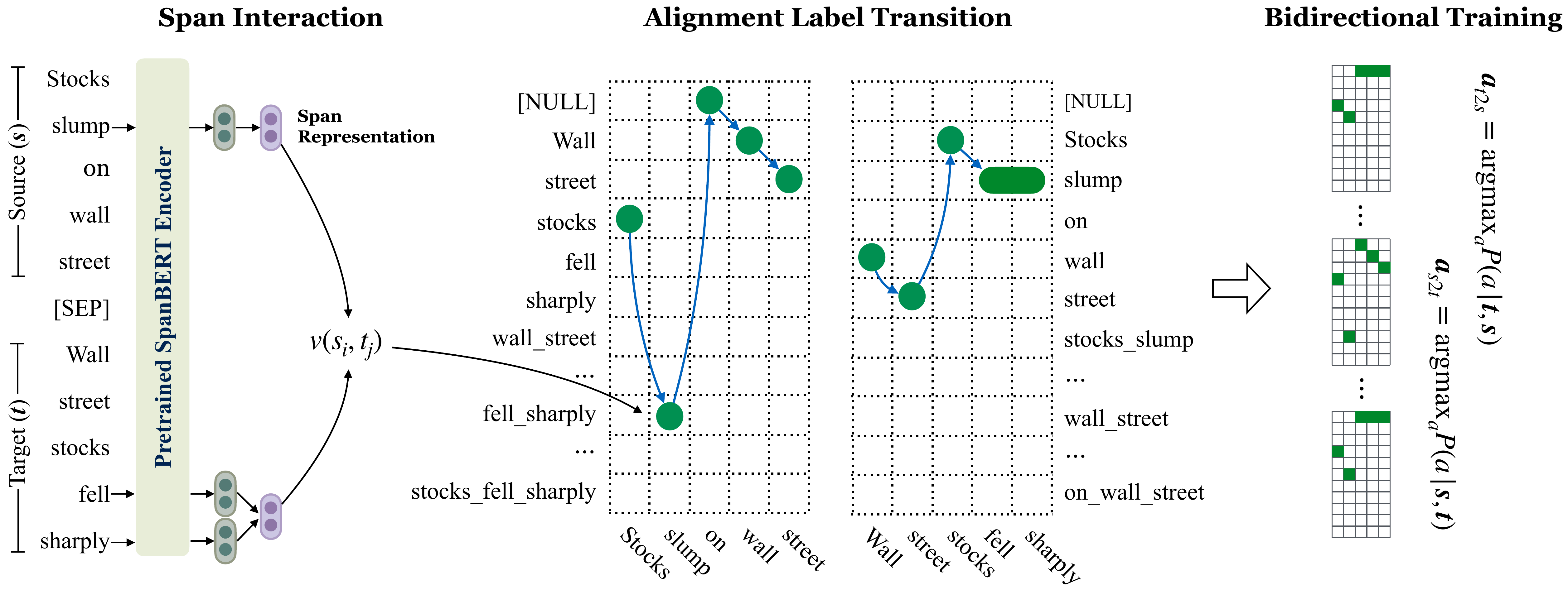}
    \caption{Illustration of our neural semi-CRF word alignment model.} 
    \label{fig:model_overview}
    \vspace{-0.65cm}
\end{figure*}

In this section, we first describe the problem formulation for monolingual word alignment, then present the architecture of our neural semi-CRF word alignment model (Figure \ref{fig:model_overview}).

\subsection{Problem Formulation}
 We formulate word alignment as a sequence tagging problem following previous works \cite{blunsom2006discriminative,yao-jacana-phrase-align2013}. Given a source sentence \(\bm{s}\) and a target sentence \(\bm{t}\) of the same language, the span alignment \(\bm{a}\) consists of a sequence of tuples \((i,j)\), which indicates that span \(s_i\) in the source sentence is aligned with span \(t_j\) in the target sentence. More specifically, \(a_{i}=j\) means source span $s_i$ is aligned with target span $t_j$. We consider all spans up to a maximum length of $D$ words. Given a source span $s_i$ of $d$ ($d \leq D$) words $[s^w_{b_i}, s^w_{b_i+1}, ..., s^w_{b_i+d-1}]$, where $b_i$ is the beginning word index, its corresponding label $a_i$ means every word within the span $s_i$ is aligned to the target span $t_{a_i}$. That is, the word-level alignments \(a^w_{b_i}, a^w_{b_i+1}, ..., a^w_{b_i+d-1}\) have the same value $j$. We use \(\bm{a^w}\) to denote the label sequence of alignments between words and $s^w_{b_i}$ to denote the $b_i$th word in the source sentence. There might be cases where span \(s_i\) is not aligned to any words in the target sentence, then \(a_{i}=\texttt{[NULL]}\). When $D\geq2$, the Markov property would no longer hold for word-level alignment labels, but for span-level labels. That is, \(a_{i}\) depends on \(a^w_{b_i-1}\), the position in the target sentence where the source span (with ending word index $b_i-1$) that precedes the current span $s_i$ is aligned to. We therefore design a discriminative model using semi-Markov conditional random fields \cite{sarawagi2005semi} to segment the source sentence and find the best span alignment, which we present below. One unique aspect of our semi-Markov CRF model is that it utilizes a varied set of labels for each sentence pair. 

\subsection{Our Model}
\label{sec:model}

The conditional probability of alignment $\bm{a}$ given a sentence pair ${\bm s}$ and ${\bm t}$  is defined as follows: 
\begin{equation}
\label{eq:1}
\resizebox{0.56\hsize}{!}{
     $ p(\bm{a}|\bm {s} , \bm t) = \frac{\displaystyle e^{\psi({\bm a}, {\bm s}, {\bm t})}}{ \sum_{{\bm a'} \in {\mathcal{A}}} \displaystyle e^{\psi({\bm a'}, {\bm s}, {\bm t})}} $ 
    }
\end{equation}
\noindent where the set $\mathcal{A}$ denotes all possible  alignments between the two sentences. The potential function \(\psi\) can be decomposed into:

\vspace{-0.5cm}

\begin{equation}
\label{eq:2}
\begin{split}
\psi({\bm a}, {\bm s}, {\bm t}) = \sum\limits_{i}{ \upsilon(s_{i},}&{t_{a_i}) + \tau( a^w_{b_i-1}, a_{i})} + \\[-5pt] & cost({\bm a}, {{\bm a}}^*)
\end{split}
\end{equation}

\noindent where ${i}$ denotes the indices of a subset of source spans that are involved in the alignment ${\bm a}$; ${\bm a}^*$ represents the gold alignment sequence at  span-level. The potential function \(\psi\) consists of three elements, of which the first two compose  negative log-likelihood loss: the span interaction function $\upsilon$, which accounts the similarity between a source span and a target span; the Markov transition function $\tau$, which models the transition of alignment labels between adjacent source spans; the $cost$ is implemented with Hamming loss to encourage the predicted alignment sequence to be consistent with gold labels. Function $\upsilon$ and $\tau$ are implemented as two neural components which we describe below.

\paragraph{Span Representation Layer.} First, source and target sentences are concatenated together and encoded by the pre-trained  \(\text{SpanBERT}\) \cite{joshi-etal-2020-spanbert} model. The hidden representations in the last layer of the encoder are extracted for each WordPiece token, then averaged to form the word representations. Following previous work \cite{joshi-etal-2020-spanbert}, the span is represented by a self-attention vector computed over the representations of each word within the span, concatenated with the Transformer output states of two endpoints.

\paragraph{Span Interaction Layer.} The semantic similarity score between source span $s_i$ and target span $t_j$ is calculated by a 2-layer feed-forward neural network $\text{FF}_{sim}$ with Parametric Relu (PReLU) \cite{he2015delving},\footnote{We also compared ReLU and GeLU, and found PReLU works slightly better.} after applying layer normalization to each span representation:
\begin{equation}
\resizebox{0.89\hsize}{!}{
    $\upsilon(s_{i},t_{j}) = \text{FF}_{sim}([{\bm{h}}^s_{i}; {\bm{h}}^t_{j}; |{\bm{h}}^s_{i}- {\bm{h}}^t_{j}|; {\bm{h}}^s_{i}\circ {\bm{h}}^t_{j}])$
    }
\end{equation}
where \([;]\) is concatenation and \(\circ\) is element-wise multiplication. We use $h^s_i$ and $h^t_j$ to denote the representation of source span $s_i$ and target span $t_j$, respectively.

\paragraph{Markov Transition Layer.}  Monolingual word alignment moves along the diagonal direction in most cases. To incorporate this intuition,  we propose a scoring function to model the transition between the adjacent alignment labels $a^w_{b_i-1}$ and $a_{i}$. The main feature we use is the distance  between the beginning index of current target span and the end index of the target span that the prior source span is aligned to. The distance is binned into 1 of 13 buckets with the following boundaries [-11, -6, -4, -3, -2, -1, 0, 1, 2, 3, 5, 10], and each bucket is encoded by a 128-dim randomly initialized embedding. It is then transformed into a real-value score by a 1-layer feed forward neural network.

\paragraph{Training and Inference.} During training, we minimizes the negative log-likelihood  of the gold alignment $\bm{a}^{*}$, and  the model is trained from both directions (source to target, target to source):
\vspace{-0.4cm}

\begin{equation}
\resizebox{0.88\hsize}{!}{
     $\sum\limits_{(\bm{s}, \bm{t}, \bm{a}^{*})}- \text{log } p(\bm{a}^{*}_{s2t}|\bm{s, t}) - \text{log } p(\bm{a}^{*}_{t2s}|\bm{t, s})$
     }
\end{equation}

\noindent where \(\bm{a}^{*}_{s2t}\) and \(\bm{a}^{*}_{t2s}\) represent the gold alignment labels from both directions. 

During inference, we use the Viterbi algorithm to find the optimal alignment. There are different strategies to merge the outputs from two directions, including intersection, union, grow-diag \cite{koehn2009statistical}, bidi-avg \cite{nagata-etal-2020-supervised}, etc.  It can be seen as a hyper-parameter and decided based on the dev set. In this work, we use intersection in our semi-CRF model for all experiments.

\subsection{Implementation Details}
We implement our model in PyTorch \cite{paszke2017automatic}. We use the Adam optimizer and set both the learning rate and weight decay as 1e-5. We set the maximum span size to 3 for our neural semi-CRF model, which can converge within 5 epochs. The neural semi-CRF model has $\sim$2 hour training time per epoch for MultiMWA-MTRef, measured on a single GeForce GTX 1080 Ti GPU.


\setlength{\tabcolsep}{2pt}
\begin{table*}[tp]
\centering
\resizebox{\textwidth}{!}{%
\renewcommand{\arraystretch}{1.2}
\begin{threeparttable}
\begin{tabular}{lcccccr@{}lr@{}lccc} 

\toprule
\textbf{Datasets} & \textbf{\#Train} & \textbf{\#Dev} & \textbf{\#Test} & \textbf{Length} &  \textbf{\%aligned} &  \textbf{\%word/} & \textbf{phrase} & \textbf{\%id/} & \textbf{non-id} &  \textbf{Genre}        & \textbf{External} & \textbf{License}   \\ 
\toprule
\multicolumn{11}{l}{\textit{\textbf{{Existing Monolingual Word Alignment Datesets}}} }                          \\ \hline

MSR RTE \cite{brockett2007aligning}                                        & 800              & --             & 800             & 29 / 11      & 37.9             & 90.0 /&\text{ }10.0              & 76.6 /&\text{ }23.4   & Misc.  & -- & Free          \\
Edinburgh++ \cite{thadani2012joint}                                    & 714              & --             & 306             & 22 / 22                         & 85.7             & 77.7 /&\text{ }22.3              & 67.2 /&\text{ }32.8    & Misc.  & -- & Free           \\ [-0.35em]
iSTS \cite{agirre2016semeval_task2}                                           & 1,506            & --             & 750             & 9 / 9      & 74.0             & 6.5 /&\text{ }93.5               & 23.3 /&\text{ }76.7 & \begin{tabular}{@{}c@{}}  News \\[-0.55em]  Image captions\end{tabular}   & Chunking & Free            \\ [-0.25em]
SPADE / ESPADA$^\dagger$ \shortcite{arase-tsujii-2018-spade, arase-tsujii-2020-compositional}                                 & 1,916            & 50             & 151             & 23 / 23                         & 81.9              & 44.0 /&\text{ }56.0                & 72.3 /&\text{ }27.7   & News   & Parsing & LDC             \\ 
\toprule
\multicolumn{11}{l}{\textit{\textbf{{Our Multi-Genre Monolingual Word Alignment (MultiMWA) Benchmark}}} }                          \\ \hline

MultiMWA-MTRef                                   & 2,398            & 800            & 800             & 22 / 17                        & 88.6             & 62.0 /&\text{ }38.0              & 52.6 /&\text{ }47.3   & News   & -- & Free            \\
MultiMWA-Wiki                                          & 2,514            & 533            & 1,052           & 30 / 29                      & 91.8             & 95.6 /&\text{ }4.4               & 94.1 /&\text{ }5.9   & Wikipedia    & -- & Free            \\
MultiMWA-Newsela                                       & --               & --             & 500             & 27 / 23                          & 76.5             & 74.6 /&\text{ }25.4              & 67.1 /&\text{ }32.9  & News     & -- & Free$^{*}$          \\
MultiMWA-arXiv                                         & --               & --             & 200             & 29 / 28                & 87.8             & 96.6 /&\text{ }3.4               & 93.4 /&\text{ }6.6     & Scientific writing & -- & Free         \\
\textbf{Total} & \textbf{4,912} & \textbf{1,333} & \textbf{2,552}  & \textbf{26} / \textbf{23} & \textbf{89.4}             & \textbf{79.3} /&\text{ }\textbf{20.7}               & \textbf{73.8} /&\text{ }\textbf{26.2}     & \textbf{all above}  & \textbf{--} & \textbf{Free} \\
\bottomrule
\end{tabular}
\end{threeparttable}
}

\caption{Statistics of our new MultiWMA benchmark and existing datasets. \textbf{Length} of the longer/shorter sentence in each pair is measured by the number of tokens.   \textbf{\%aligned} is the percentage of aligned words among all words.  \textbf{\%word/phrase} denotes the percentage of word alignment and phrasal alignment. \textbf{\%id/non-id} specifies the percentage of identical (e.g., \textit{Lloyd} $\leftrightarrow$ \textit{Lloyd}) and non-identical (e.g., \textit{conduct} $\leftrightarrow$ \textit{performed}) alignments. \textbf{Externel} indicates whether the annotation relies on additional linguistic information.  $^\dagger$ESPADA (train) has not been released at the time of writing; statistics are based on the SPADE (dev/test) dataset. $^{*}$Newsela data is free for academic research but license needs to be requested at: \url{https://newsela.com/data}.}
\label{statistics_on_datasets}
  \vspace{-0.8cm}
\end{table*}

\section{A Multi-Genre Benchmark for Monolingual Word Alignment}
\label{sec:benchmark}

In this section, we present the manually annotated \textbf{Multi}-genre \textbf{M}onolingual \textbf{W}ord \textbf{A}lignment (MultiMWA) benchmark that consists of  four datasets of different text genres. As summarized in Table \ref{statistics_on_datasets}, our new benchmark is the largest to date and of higher quality compared to existing datasets. In contrast to iSTS \cite{agirre2016semeval_task2} and SPADE/ESPADA \cite{arase-tsujii-2018-spade, arase-tsujii-2020-compositional}, our annotation does not rely on external chunking or parsing that may introduce errors or restrict the granularity and flexibility. Our benchmark contains both token alignments and a significant portion of phrase alignments as they are semantically equivalent as a whole.
Our benchmark also contains a large portion of semantically similar but not strictly equivalent sentence pairs, which are common in text-to-text generation tasks and thus important for evaluating the monolingual word alignment models under this realistic setting. 

For all four datasets, we closely follow the standard 6-page annotation guideline\footnote{\url{http://www.cs.jhu.edu/~ccb/publications/paraphrase_guidelines.pdf}} from \cite{callison2006annotation} and further extend it to improve the phrase-level annotation consistency (more details in Appendix \ref{sec:update_guideline}). We describe each of the four datasets below.

\paragraph{MultiMWA-MTRef.} We create this dataset by annotating 3,998 sentence pairs from the MTReference \cite{yao-phd-2014}, which are human references used in a machine translation task. The original labels in MTReference were annotated by crowd-sourcing workers on Amazon Mechanical Turk following the guideline from \cite{callison2006annotation}. In an early pilot study, we discovered that these crowd-sourced annotations are noisy and contain many obvious errors. It only gets  73.6/96.3/83.4 for Precision/Recall/F$_1$ on a random sample of 100 sentence pairs, when compared to the labels we manually corrected.

To address the lack of reliable annotation, we hire two in-house annotators to correct the original labels using GoldAlign\footnote{\url{https://github.com/ajdagokcen/goldalign-repo}} \cite{GOKCEN16.679}, an annotation tool for monolingual word alignment. Both annotators have  linguistic background and extensive NLP annotation experience. We provide a three-hour training session to the the annotators, during which they are asked to align 50 sentence pairs and discuss until consensus. Following previous work, we calculate the inter-annotator agreement as 84.2 of F$_1$ score for token-level non-identical alignments by comparing one annotator's annotation against the other's. The alignments between identical words are usually easy for human annotators. After merging the the labels from both annotators, we create a new split of 2398/800/800 for train/dev/test set. To ensure the quality, an adjudicator further exams the dev and test sets and constructs the final labels.

\paragraph{MultiMWA-Newsela.}  Newsela corpus \cite{xu-etal-2015-problems} consists of 1,932 English news articles and their simplified versions written by professional editors. It has been widely used in text simplification research \cite{xu-etal-2016-optimizing, zhang-lapata-2017-sentence,zhong2020discourse}. We randomly select 500 complex-simple sentence pairs from the test set of Newsela-Auto \cite{jiang2020neural},\footnote{More specifically, we sample from the exact test set used in Table 2 in \newcite{maddela2020controllable}.} which is the newest sentence-aligned version of Newsela. 214 of these 500 pairs contain sentence splitting. An in-house annotator\footnote{This annotator has annotated MultiMWA-MTRef.} labels the word alignment by correcting the outputs from GIZA++ \cite{och03:asc}.

\paragraph{MultiMWA-arXiv.} The arXiv\footnote{\url{https://arxiv.org/}} is an open-access platform that stores more than 1.7 million research papers with their historical versions. It has been used to study paraphrase generation \cite{dong2021parasci} and statement strength \cite{tan-lee-2014-corpus}. We first download the \LaTeX \text{} source code for 750 randomly sampled papers and their historical versions, then use OpenDetex\footnote{\url{https://github.com/pkubowicz/opendetex}} package to extract plain text from them. We use the trained neural CRF sentence alignment model \cite{jiang2020neural} to align sentences between different versions of the papers and sample  200 non-identical aligned sentence pairs for further annotation. The word alignment is annotated in a similar procedure to that of the MultiMWA-Wiki.

\paragraph{MultiMWA-Wiki.} Wikipedia has been widely used in text-to-text tasks, including text simplification \cite{jiang2020neural}, sentence splitting \cite{botha-etal-2018-learning}, and neutralizing bias language \cite{pryzant2020automatically}. We follow the method in \cite{pryzant2020automatically} to extract parallel sentences from Wikipedia revision history dump (dated 01/01/2021) and randomly sample 4,099 sentence pairs for further annotation. We first use an earlier version of our neural semi-CRF word aligner (\S \ref{sec:neuralcrf}) to automatically align words for the sentence pairs, then ask two in-house annotators to correct the aligner's outputs. The inter-annotator agreement is 98.1 at token-level measured by F$_1$.\footnote{The inter-annotator agreement is much higher compared to that of MultiMWA-MTRef, as the parallel sentences extracted from Wikipedia revision history have more overlap.}  We split the data into 2514/533/1052 sentence pairs for train/dev/test sets.

\begin{table*}[!ht]
\small
\centering
\resizebox{\textwidth}{!}{%
\renewcommand{\arraystretch}{1.25}
\begin{tabular}{ l@{\hspace{2.5ex}}ccc@{\hspace{1.6ex}}c|ccc@{\hspace{1.6ex}}c|ccc@{\hspace{1.6ex}}c } 
\toprule
 \\[-1em]
 \multicolumn{1}{l}{\multirow{3}{*}{\large \bf{Models}}} & \multicolumn{4}{c}{\large\textbf{MultiMWA-MTRef$_{Sure}$}} & \multicolumn{4}{c} {\large\textbf{MultiMWA-MTRef$_{Sure+Poss}$}} & \multicolumn{4}{c}{\large\bf{MultiMWA-Wiki}} \\ \rule{0pt}{3ex}
  & \doubleRowCell{\bf P}{$P_i$}{$P_n$} & \doubleRowCell{\bf R}{$R_i$}{$R_n$} & \doubleRowCell{\bf F$_1$}{F$_1$$_i$}{F$_1$$_n$} & \fontsize{10}{11}\selectfont \bf{EM} & \doubleRowCell{\bf P}{$P_i$}{$P_n$} & \doubleRowCell{\bf R}{$R_i$}{$R_n$} & \doubleRowCell{\bf F$_1$}{F$_1$$_i$}{F$_1$$_n$} & \fontsize{10}{11}\selectfont \bf{EM} & \doubleRowCell{\bf P}{$P_i$}{$P_n$} & \doubleRowCell{\bf R}{$R_i$}{$R_n$} & \doubleRowCell{\bf F$_1$}{F$_1$$_i$}{F$_1$$_n$} & \fontsize{10}{11}\selectfont \bf{EM} \\  \midrule
\fontsize{12}{12}\selectfont  JacanaToken \cite{yao-jacana-wordalign-acl2013} & \doubleRowCell{87.9}{94.4}{65.1} & \doubleRowCell{72.2}{94.7}{41.3} & \doubleRowCell{79.3}{94.6}{50.5} & \fontsize{11}{11}\selectfont 2.6 & \doubleRowCell{82.8}{93.3}{61.7} & \doubleRowCell{70.5}{96.7}{43.6} & \doubleRowCell{76.2}{95.0}{51.1} & \fontsize{11}{11}\selectfont 1.3 & \doubleRowCell{98.8}{99.3}{77.1} & \doubleRowCell{95.7}{99.5}{71.6} & \doubleRowCell{97.2}{99.4}{74.3} & \fontsize{11}{11}\selectfont 59.8 \\ 
 
\fontsize{12}{12}\selectfont   JacanaPhrase \cite{yao-jacana-phrase-align2013} & \doubleRowCell{84.4}{94.1}{58.5} & \doubleRowCell{72.4}{95.3}{40.7} & \doubleRowCell{78.0}{94.7}{48.0} &  \fontsize{11}{11}\selectfont 1.9 & \doubleRowCell{82.8}{93.3}{61.4} & \doubleRowCell{70.0}{96.2}{42.5} & \doubleRowCell{75.8}{94.8}{50.3} & \fontsize{11}{11}\selectfont 1.4 & \doubleRowCell{92.8}{98.5}{44.4} & \doubleRowCell{97.0}{99.8}{49.1} & \doubleRowCell{94.9}{99.2}{46.6} & \fontsize{11}{11}\selectfont 27.4 \\ 

\fontsize{12}{12}\selectfont   PipelineAligner \cite{sultan2014back}  & \doubleRowCell{\textbf{96.0}}{98.1}{78.9} & \doubleRowCell{67.7}{93.3}{30.6} & \doubleRowCell{79.4}{95.6}{44.1} & \fontsize{11}{11}\selectfont 2.5 & \doubleRowCell{\textbf{97.1}}{98.3}{82.9} & \doubleRowCell{60.8}{92.9}{23.9} & \doubleRowCell{74.8}{95.5}{37.1} & \fontsize{11}{11}\selectfont 1.0 & \doubleRowCell{\textbf{99.5}}{99.6}{66.2} & \doubleRowCell{94.9}{99.6}{60.0} & \doubleRowCell{97.1}{99.6}{62.9} & \fontsize{11}{11}\selectfont 53.4 \\ \cmidrule{1-13}

\fontsize{12}{12}\selectfont   QA-based Aligner  & \doubleRowCell{88.4}{98.2}{76.3} & \doubleRowCell{\textbf{92.3}}{99.2}{83.9} & \doubleRowCell{90.3}{98.7}{79.9} & \fontsize{11}{11}\selectfont 14.0 & \doubleRowCell{91.3}{98.5}{84.1} & \doubleRowCell{\textbf{92.9}}{99.2}{86.9} & \doubleRowCell{92.1}{98.9}{85.5} & \fontsize{11}{11}\selectfont 21.3 & \doubleRowCell{97.4}{99.5}{82.3} & \doubleRowCell{\textbf{97.9}}{99.8}{81.9} & \doubleRowCell{\textbf{97.6}}{99.7}{82.1} & \fontsize{11}{11}\selectfont 67.4 \\

 \fontsize{12}{12}\selectfont   Neural CRF Aligner  & \doubleRowCell{87.6}{97.3}{74.2} & \doubleRowCell{91.6}{99.5}{82.2} & \doubleRowCell{89.5}{98.4}{78.0} & \fontsize{11}{11}\selectfont 10.8 & \doubleRowCell{91.5}{98.5}{83.4} & \doubleRowCell{90.2}{99.2}{82.1} & \doubleRowCell{90.8}{98.8}{82.7} & \fontsize{11}{11}\selectfont 16.9 &  \doubleRowCell{96.5}{99.3}{80.6} & \doubleRowCell{97.6}{99.6}{80.6} & \doubleRowCell{97.1}{99.4}{80.6} & \fontsize{11}{11}\selectfont 63.5 \\ 
  
 \fontsize{12}{12}\selectfont    Neural semi-CRF Aligner  & \doubleRowCell{90.6}{98.9}{78.9} & \doubleRowCell{{90.3}}{98.9}{79.1} & \doubleRowCell{\textbf{90.5}}{98.9}{79.0} & \fontsize{11}{11}\selectfont \textbf{14.1} & \doubleRowCell{94.7}{99.3}{89.1} & \doubleRowCell{90.2}{98.7}{82.3} & \doubleRowCell{\textbf{92.4}}{99.0}{85.5} & \fontsize{11}{11}\selectfont \textbf{23.3}&  \doubleRowCell{{97.7}}{99.6}{82.8} & \doubleRowCell{97.5}{99.7}{80.8} & \doubleRowCell{\textbf{97.6}}{99.7}{81.8$^*$} & \fontsize{11}{11}\selectfont \textbf{68.5} \\ 
\bottomrule
\end{tabular}
}
\caption{\label{results_in_domain} In-domain evaluation of different monolingual word alignment models on the MultiMWA benchmark. We report the precision (\textbf{P}), recall (\textbf{R}), \textbf{F$_1$}, and exact match (\textbf{EM}), which is the percentage of sentence pairs for which model predictions are exactly same as gold labels for the entire sentence. For each metric, we also report the performance on identical alignments (P{$_i$}, R{$_i$}, F$_1${$_i$}) and non-identical alignments (P{$_n$}, R{$_n$}, F$_1${$_n$}) separately. $^*$ MultiMWA-Wiki contains only about 5\% non-identical alignment.}
\vspace{-0.5cm}
\end{table*}

\begin{table*}[!ht]
\small
\centering
\resizebox{\textwidth}{!}{%
\renewcommand{\arraystretch}{1.25}
\begin{tabular}{ l@{\hspace{2.4ex}}ccc@{\hspace{1.6ex}}c|ccc@{\hspace{1.6ex}}c|ccc@{\hspace{1.6ex}}c } 
\toprule 
 \\[-1em]
 \multicolumn{1}{l}{\multirow{3}{*}{\large \bf{Models}}} & \multicolumn{4}{c}{\large\textbf{MultiMWA-Newsela}} & \multicolumn{4}{c} {\large\textbf{MultiMWA-arXiv}} & \multicolumn{4}{c}{\large\bf{MultiMWA-Wiki}} \\ \rule{0pt}{3ex}
  & \doubleRowCell{\bf P}{$P_i$}{$P_n$} & \doubleRowCell{\bf R}{$R_i$}{$R_n$} & \doubleRowCell{\bf F$_1$}{F$_1$$_i$}{F$_1$$_n$} & \fontsize{10}{11}\selectfont \bf{EM} & \doubleRowCell{\bf P}{$P_i$}{$P_n$} & \doubleRowCell{\bf R}{$R_i$}{$R_n$} & \doubleRowCell{\bf F$_1$}{F$_1$$_i$}{F$_1$$_n$} & \fontsize{10}{11}\selectfont \bf{EM} & \doubleRowCell{\bf P}{$P_i$}{$P_n$} & \doubleRowCell{\bf R}{$R_i$}{$R_n$} & \doubleRowCell{\bf F$_1$}{F$_1$$_i$}{F$_1$$_n$} & \fontsize{10}{11}\selectfont \bf{EM} \\ 
 \hline \rule{-1.7pt}{3.7ex}
\fontsize{12}{12}\selectfont JacanaToken \cite{yao-jacana-wordalign-acl2013} & \doubleRowCell{85.5}{91.2}{60.1} & \doubleRowCell{74.9}{97.5}{39.7} & \doubleRowCell{79.8}{94.3}{47.9} & \fontsize{11}{11}\selectfont 11.0 & \doubleRowCell{94.9}{97.3}{72.6} & \doubleRowCell{96.8}{99.5}{73.4} & \doubleRowCell{95.8}{98.4}{73.0} & \fontsize{11}{11}\selectfont 49.0 & \doubleRowCell{94.7}{98.4}{51.2} & \doubleRowCell{96.9}{99.9}{50.1} & \doubleRowCell{95.8}{99.2}{50.6} & \fontsize{11}{11}\selectfont 33.3 \\ 
 
\fontsize{12}{12}\selectfont   JacanaPhrase \cite{yao-jacana-phrase-align2013} & \doubleRowCell{84.3}{91.3}{53.9} & \doubleRowCell{75.0}{97.4}{38.6} & \doubleRowCell{79.4}{94.3}{45.0} &  \fontsize{11}{11}\selectfont 8.2 & \doubleRowCell{90.9}{97.1}{53.2} & \doubleRowCell{96.6}{99.1}{64.7} & \doubleRowCell{93.7}{98.1}{58.4} & \fontsize{11}{11}\selectfont 31.5 & \doubleRowCell{92.9}{98.5}{44.9} & \doubleRowCell{96.9}{99.8}{49.6} & \doubleRowCell{94.9}{99.1}{47.1} & \fontsize{11}{11}\selectfont  28.0\\ 

\fontsize{12}{12}\selectfont   PipelineAligner \cite{sultan2014back}  & \doubleRowCell{\textbf{95.2}}{96.9}{64.4} & \doubleRowCell{69.4}{95.3}{25.4} & \doubleRowCell{80.3}{96.1}{36.5} & \fontsize{11}{11}\selectfont 10.0 & \doubleRowCell{\textbf{98.5}}{98.8}{68.3} & \doubleRowCell{94.6}{99.0}{62.4} & \doubleRowCell{96.5}{98.9}{65.2} & \fontsize{11}{11}\selectfont 49.0 & \doubleRowCell{\textbf{99.5}}{99.6}{66.2} & \doubleRowCell{94.9}{99.6}{60.0} & \doubleRowCell{97.1}{99.6}{62.9} & \fontsize{11}{11}\selectfont 53.4 \\

 \hline \rule{-1.7pt}{3.7ex}
\fontsize{12}{12}\selectfont   QA-based Aligner & \doubleRowCell{84.8}{95.3}{69.4} & \doubleRowCell{\textbf{87.9}}{99.1}{71.4} & \doubleRowCell{86.2}{97.1}{70.4} & \fontsize{11}{11}\selectfont 16.2 & \doubleRowCell{93.9}{98.0}{70.7} & \doubleRowCell{94.3}{95.0}{79.9} & \doubleRowCell{94.1}{96.5}{75.0} & \fontsize{11}{11}\selectfont 27.0 & \doubleRowCell{96.1}{99.3}{76.2} & \doubleRowCell{\textbf{98.2}}{99.8}{78.3} & \doubleRowCell{97.2}{99.5}{77.3} & \fontsize{11}{11}\selectfont 57.8 \\

\fontsize{12}{12}\selectfont   Neural CRF Aligner & \doubleRowCell{88.2}{95.3}{72.3} & \doubleRowCell{85.0}{99.0}{66.3} & \doubleRowCell{86.6}{97.1}{69.1} & \fontsize{11}{11}\selectfont 15.6 & \doubleRowCell{92.9}{96.4}{62.9} & \doubleRowCell{\textbf{98.7}}{99.8}{73.3} & \doubleRowCell{95.7}{98.0}{67.7} & \fontsize{11}{11}\selectfont 43.5 & \doubleRowCell{96.1}{99.1}{70.5} & \doubleRowCell{98.0}{99.9}{71.9} & \doubleRowCell{97.0}{94.5}{71.2} & \fontsize{11}{11}\selectfont 52.1 \\ 
  
 \fontsize{12}{12}\selectfont    Neural semi-CRF Aligner & \doubleRowCell{89.4}{96.7}{76.1} & \doubleRowCell{85.0}{98.4}{66.5} & \doubleRowCell{\textbf{87.2}}{97.6}{71.0} & \fontsize{11}{11}\selectfont \textbf{21.6} & \doubleRowCell{96.2}{98.9}{79.3} & \doubleRowCell{98.4}{99.6}{83.0} & \doubleRowCell{\textbf{97.3}}{99.3}{81.1} & \fontsize{11}{11}\selectfont \textbf{62.5} & \doubleRowCell{97.2}{99.6}{80.4} & \doubleRowCell{97.6}{99.5}{79.5} & \doubleRowCell{\textbf{97.4}}{99.5}{79.9} & \fontsize{11}{11}\selectfont \textbf{64.8} \\
\bottomrule
\end{tabular}
}
\caption{\label{results_out_of_domain} Out-of-domain evaluation of different monolingual word alignment models on the MultiMWA benchmark. All the models in this table are trained on the {MultiMWA-MTRef}\(_{Sure+Poss}\) dataset.}
\vspace{-0.8cm}
\end{table*}

\section{Experiments}
In this section, we present both in-domain and out-of-domain evaluations for different word alignment models on our MultiWMA benchmark. We also provide a detailed error analysis of our neural semi-CRF model and an ablation study to analyze the importance of each component.

\subsection{Baselines}
\label{sec:baseline}
We introduce a novel state-of-the-art baseline by adapting the \textbf{QA-based method} in \cite{nagata-etal-2020-supervised}, which has not previously applied to monolingual word alignment but only bilingual word alignment. This method treats the word alignment problem as a collection of independent predictions from every token in the source sentence to a span in the target sentence, which is then solved by fine-tuning multilingual BERT \cite{devlin2018bert} similarly as for SQuAD-style question answering task. Taking the sentence pair in Figure 1 as an example, the word to be aligned is marked by $\P$ in the source sentence and concatenated with the entire target sentence to form the input as ``\textit{With Canadian $\cdots$ $\P$conduct$\P$ $\cdots$ his model. Lkoyd performed $\cdots$ his model. }''. A span prediction model based on fine-tuning multilingual BERT is then expected to extract \textit{performed} from the target sentence. The predictions from both directions (source to target, target to source) are symmetrized to produce the final alignment, using a probability threshold of 0.4 instead of the typical 0.5.

We change to use standard BERT in this model for monolingual alignment and find that the 0.4 threshold chosen by \citet{nagata-etal-2020-supervised} is almost optimal in maximizing the F$_1$ score on our MultiMWA-MTRef dataset. This QA-based method alone outperforms all existing models for monolingual word alignment, including: \textbf{JacanaToken} aligner \cite{yao-jacana-wordalign-acl2013}, which is a CRF model using hand-crafted features and external resources; \textbf{JacanaPhrase} aligner \cite{yao-jacana-phrase-align2013}, which is a semi-CRF model relying on feature templates and external resources; \textbf{PipelineAligner} \cite{sultan2014back}, which is a pipeline system that utilizes word similarity and contextual information with heuristic algorithms. We also create a variation of our model, a \textbf{Neural CRF aligner}, in which all modules remain the same but the max span length is set to 1, to evaluate the benefits of span-based alignments.

\subsection{Experimental Results}
Following the literature \cite{thadani2012joint,yao-jacana-wordalign-acl2013,yao-jacana-phrase-align2013}, we present results under both $Sure$ and $Sure+Poss$ settings for the MultiMWA-MTRef dataset. $Sure+Poss$ setting includes all the annotated alignments, and $Sure$ only contains a subset of them which are agreed by multiple annotators. We consider $Sure+Poss$ as the default setting for all the other three datasets.

The in-domain evaluation results are shown in Table \ref{results_in_domain}. The neural models are working remarkably well in comparison to the non-neural methods, especially as measured by Exact Matches (EM). On both MTRef and Wiki datasets, our neural semi-CRF model achieves the best  F$_1$ and EM. QA-based aligner also has competitive performance with strong recall, however, its precision is lower compared to our model. It is worthy to note that our model has a modular design, and can be more easily adjusted than QA-based method to suit different datasets and downstream tasks.

Table \ref{results_out_of_domain} presents the out-of-domain evaluation results. Our neural models achieve the best performance across all three datasets. This demonstrates the generalization ability of our model, which can be useful in the downstream applications.

\subsection{Ablation Study}

Table \ref{ablation_study} shows the ablation study for our neural semi-CRF model. $F_1$ and EM drops by 1.3 and 4.4 points respectively after replacing SpanBERT with BERT, indicating the importance of optimized pre-trained representations. Markov transition layer contributes mainly to the alignment accuracy (EM).  We have experimented with different strategies to merge the outputs from two directions: intersection yields better precision, grow-diag and union bias towards recall. Leveraging the span
interaction matrix generated by our model (details in \S\ref{sec:model}), we design a simple post-processing rule to extend the phrasal alignment to spans that are longer than 3 tokens. Adjacent target words are gradually included if they have very high semantic similarity with the same source span. This rule further improves recall and achieves the best F$_1$ on the MultiMWA-MTRef.

\setlength{\tabcolsep}{4pt}
\begin{table}[!t]
\small
\centering
\resizebox{0.9\linewidth}{!}{%
\renewcommand{\arraystretch}{1.4}
\begin{tabular}{lccc}
\toprule
Neural semi-CRF Aligner  & $\text{F}_\text{1}$ & $\text{EM}$ & $\Delta_{\text{F}_\text{1}}$/$\Delta_{\text{EM}}$ \\
\toprule
w/ SpanBERT & 92.1 & 23.3 & \multicolumn{1}{c}{0.0 / 0.0} \\

w/ BERT & 90.8 & 18.9 & -1.3 / -4.4\\ \hline
w/o Transition Layer & 91.9 & 21.3 & -0.2 / -2.0 \\ \hline
w/ post-processing & 92.1 & 23.3  & \multicolumn{1}{c}{0.0 / 0.0} \\
w/ intersection & 92.0 & 21.8 & -0.1/ -1.5 \\
w/ union & 91.1 & 20.1 & -1.0 / -3.2 \\
w/ grow-diag & 91.5 & 20.6 & -0.6 / -2.7 \\

\bottomrule
\end{tabular}
}
\caption{\label{ablation_study} Ablation study of our neural semi-CRF aligner with each component removed or swapped. The results are based on the dev set of MTRef\(_{Sure+Poss}\). }
\vspace{-0.5cm}
\end{table}
\setlength{\tabcolsep}{3pt}

\subsection{Error Analysis}
\label{section_error_analysis}

We sample 50 sentence pairs from the dev set of MultiMWA-MTRef and analyze the errors under \textbf{Sure+Poss} setup.\footnote{The strict \textbf{Sure only} labels exclude many alignments that are critical for certain applications, such as label projection. We thus focus on the \textbf{Sure+Poss} labels for error analysis.} Figure \ref{error_analysis} shows how the performance of different alignment models would improve, if we resolve each of the 7 types of errors. We discuss the categorization of errors  and their breakdown percentages below:

\paragraph{Phrase Boundary (58.6\%).} The phrase boundary error (see \textcircled{3} in Figure \ref{fig:example} for an example) is the most prominent error in all models, attributing 7.6 points of F$_1$ for JacanaPhrase, 5.7 for QA aligner, and 4.7 for neural semi-CRF aligner. For another example, instead of 3x2 alignment \(\textit{funds for research}\leftrightarrow \textit{research funding}\), our model captures two 1x1 alignments,   \(\textit{funds}\leftrightarrow \textit{funding}\) and \(\textit{research}\leftrightarrow \textit{research}\). This is largely due to the fact that alignments are not limited to linguistic phrases (e.g., noun phrases, verb phrases, etc.), but rather, include non-linguistic phrases. It could also be challenging to handle longer spans, such as \(\textit{keep his position}\leftrightarrow \textit{protect himself from being removed}\) (more on this in Appendix \ref{sec:alignment_shape}). Although we use SpanBERT for better phrase representation, there is still room for improvement.

\paragraph{Function Words (19.1\%).} Function words can be tricky to align when rewording and reordering happens, such as \textcircled{2}. Adding on the complexity, same function word may appear more than once in one sentence. This type of error is common in all the models we experiment with. It attributes 4.7 points of F$_1$ for JacanaPhrase, 1.3 for QA aligner, and 1.5 for our neural semi-CRF aligner.

\paragraph{Content Words (14.2\%).} Similar to function words, content words (e.g.,  \(\textit{security bureau}\leftrightarrow \textit{defense ministry}\)) can also be falsely aligned or missed, but the difference between neural and non-neural model is much more significant. This error type attributes 7.7 points of F$_1$ score for Jacana aligner, but only 1.1 and 0.8 for neural semi-CRF aligner and QA aligner, respectively.

\paragraph{Context Implication (5.6\%).} Some words or phrases that are not strictly semantically equivalent can also be aligned if they appear in a similar context. For example, given the source sentence \textit{`Gaza international airport was put into operation the day before'} and the target sentence \textit{`The airport began operations one day before'}, the phrase pair \(\textit{was put into}\leftrightarrow \textit{began}\) can be aligned. This type is related to 2.8 F$_1$ score improvement for Jacana aligner, but only 0.4 and 0.2 for neural semi-CRF and QA-based aligners, respectively. 

\paragraph{Debatable Labels (1.9\%).} Word alignment annotation can be subjective sometimes. Take phrase alignment \(\textit{two days of}\leftrightarrow \textit{a two-day}\) for example, it can go either way to include the function word `\textit{a}' in the alignment, or not.

\paragraph{Name Variations (0.6\%).} While our neural semi-CRF model is designed to handle spelling variations or name abbreviations, it fails sometimes as shown by \textcircled{1} in Figure \ref{fig:example} as an example. Some cases can be very difficult, such as \textit{SAWS} $\leftrightarrow$ \textit{the state's supervision and control bureau of safe production}, where \textit{SAWS} stands for \textit{State Administration of Work Safety}.

\paragraph{Skip Alignment (0.0\%).} Non-contiguous tokens can be aligned to the same target token or phrase (e.g., \(\textit{owes ... to}\leftrightarrow \textit{is a result of}\)), posing a challenging situation for monolingual word aligners. However, this error is rare, as only 0.6\% of all alignments in MTRef dev set are discontinuous.

\begin{figure}[!t]
    \centering
    \begin{tikzpicture}[thick,scale=0.34,baseline]
  \small
        \fill[black]  (0,8) rectangle (1,9);
        \fill[black]  (1,8) rectangle (2,9);
        
        
        \fill[pattern=slant lines, slope=-0.7,pattern color=red]  (2,8) rectangle (3,9);
        
        \fill[black]  (0,7) rectangle (1,8);
        \fill[black]  (1,7) rectangle (2,8);
        
        \fill[pattern=slant lines, slope=-0.7,pattern color=red]  (2,7) rectangle (3,8);
        
        \fill[pattern=slant lines, slope=0.7,pattern color=blue] (2,9) rectangle (3,10);
        
        
        \fill[black]  (3,10) rectangle (4,11);
        
        \fill[pattern=slant lines, slope=0.7,pattern color=blue]  (4,10) rectangle (5,11);
        
        \fill[black]  (10,4) rectangle (12,5);
        
        \fill[black]  (12,4) rectangle (15,5);
        
        \fill[black]  (8,1) rectangle (9,2);
        
        \fill[black]  (8,2) rectangle (9,4);
        
        \fill[black]  (15,0) rectangle (16,1);
        \fill[black]  (6,6) rectangle (7,7);
        \fill[black]  (7,5) rectangle (8,6);

        \draw (0,0.5) node[left,align=right]{.};
        \draw (0,1.5) node[left,align=right]{strained};
        \draw (0,2.5) node[left,align=right]{being};
        \draw (0,3.5) node[left,align=right]{relations};
        \draw (0,4.5) node[left,align=right]{China-US};
        \draw (0,5.5) node[left,align=right]{made};
        \draw (0,6.5) node[left,align=right]{once};
        \draw (0,7.5) node[left,align=right]{detained};
        \draw (0,8.5) node[left,align=right]{being};
        \draw (0,9.5) node[left,align=right]{'s};
        \draw (0,10.5) node[left,align=right]{Fang};

        \draw (0.3,-0.1) node[right, align=left,rotate=-50]{{The}};
        \draw (1.3,-0.1) node[right, align=left,rotate=-50]{{arrest}};
        \draw (2.3,-0.1) node[right, align=left,rotate=-50]{{of}};
        \draw (3.3,-0.1) node[right, align=left,rotate=-50]{{Fang}};
        \draw (4.3,-0.1) node[right, align=left,rotate=-50]{{Fuming}};
        \draw (5.3,-0.1) node[right, align=left,rotate=-50]{{has}};
        \draw (6.3,-0.1) node[right, align=left,rotate=-50]{{temporarily}};
        \draw (7.3,-0.1) node[right, align=left,rotate=-50]{{caused}};
        \draw (8.3,-0.1) node[right, align=left,rotate=-50]{{tensions}};
        \draw (9.3,-0.1) node[right, align=left,rotate=-50]{{between}};
        \draw (10.3,-0.1) node[right, align=left,rotate=-50]{{China}};
        \draw (11.3,-0.1) node[right, align=left,rotate=-50]{{and}};
        \draw (12.3,-0.1) node[right, align=left,rotate=-50]{{the}};
        \draw (13.3,-0.1) node[right, align=left,rotate=-50]{{United}};
        \draw (14.3,-0.1) node[right, align=left,rotate=-50]{{States}};
        \draw (15.3,-0.1) node[right, align=left,rotate=-50]{{.}};

        \draw[thick] (0,0) grid (16, 11);
        
        \filldraw[color=blue!60, fill=white!80, thick](5.8,11.0) circle (0.48); \draw (6.38,11.0) node[left,align=center]{1};
        \filldraw[color=blue!60, fill=white!80, thick](1.2,10.0) circle (0.48); \draw (1.78,10.0) node[left,align=center]{2};
        \filldraw[color=red!60, fill=white!80, thick](3.7,8.0) circle (0.48); \draw (4.28,8.0) node[left,align=center]{3};
        
        \end{tikzpicture}
        \caption{Error examples of the semi-CRF word alignment model on MTRef data. Black-filled boxes denote true positives, boxes filled with blue diagonal lines are false negatives, and red slant lines are false positives. }
 \label{fig:example}
\end{figure}

\begin{figure}[!t]
    \centering
    \includegraphics[width=0.48\textwidth]{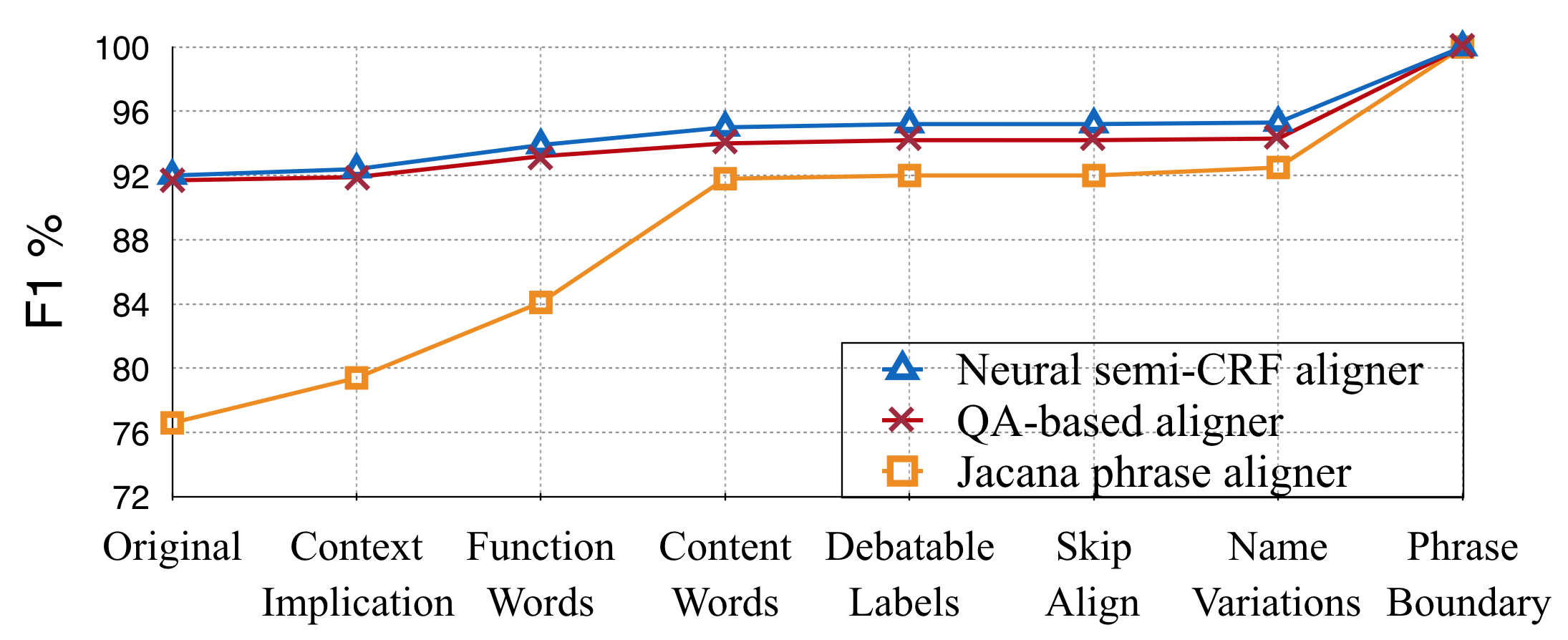}
    \caption{Performance comparison on MTRef dev set for 3 different aligners after resolving each error type.}
    \label{error_analysis}
\vspace{-0.8cm}
\end{figure}

\begin{table*}[!ht]
\small
\centering
\resizebox{.95\textwidth}{!}{%
\renewcommand{\arraystretch}{1.3}
\begin{tabular}{ llcccccccccccc } 
\toprule
\textbf{Datasets} & \textbf{Models} & \bf SARI & \bf add & \bf keep & \bf del & \bf FK & \bf SLen & \bf OLen & \bf CR & \bf \%split & \bf s-BL &\bf \%new & \bf \%eq \\
\hline
\multirow{4}{*}{Newsela-auto} & Complex (input) & 11.8 & 0.0 & 35.5 & 0.0 & 12.3 & 24.8 & 24.8 & 1.0 & 2.0 & 100.0 & 0.0 & 100.0 \\
& Simple (reference) & 86.9 & 84.7 & 78.4 & 97.6 & 6.5 & 13.3 & 13.3 & 0.63 & 0.8 & 25.7 & 33.5 & 0.0 \\
\cline{2-14}
& EditNTS & 36.6 & 1.1 & 32.9 & 75.7 & 7.5 & 14.3 & 14.3 & 0.66 & 2.4 & 50.2 & 6.5 & 1.2  \\
& EditNTS + Aligner &  \textbf{37.5} &  \textbf{1.3} &  \textbf{33.4} &  \textbf{77.9} &  7.2 & 14.3 & 14.3 & 0.66 & 1.5 & 49.1 & 7.6 & 0.8  \\
\hline
\multirow{4}{*}{Wikipedia-auto} & Complex (input) & 24.9 & 0.0 & 74.6 & 0.0 & 13.4 & 22.6 & 22.6 & 1.0 & 0.8 & 100.0 & 0.0 & 100.0  \\
& Simple (reference) & 81.7 & 66.2 & 97.5 & 81.5 & 12.2 & 21.7 & 21.7 & 0.97 & 5.4 & 64.0 & 14.8 & 16.2 \\
\cline{2-14}
& EditNTS & 36.8 & \textbf{2.1} & 68.4 & 39.8 & 12.8 & 23.6 & 23.6 & 1.06 & 1.7 & 69.7 & 12.4 & 0.6 \\
& EditNTS + Aligner  & \textbf{37.4} & 1.9 & \textbf{69.5} & \textbf{40.9} & 12.7 & 23.6 & 23.6 & 1.05 & 0.6 & 74.4 & 10.2 & 2.8 \\
\bottomrule
\end{tabular}
}
\caption{\label{results_text_simplification} Downstream application on text simplification. By incorporating our monolingual word aligner into the EditNTS \cite{dong-etal-2019-editnts} model, we improve the performance measured by \textbf{SARI} score (the main automatic metric for simplification) and its three parts: precision for delete (\textbf{del}), F$_1$ scores for \textbf{add} and \textbf{keep} operations.}
\vspace{-0.3cm}
\end{table*}

\begin{table*}[!ht]
\small
\centering
\resizebox{\textwidth}{!}{%
\renewcommand{\arraystretch}{1.2}
\begin{tabular}{ lcccccccccccc } 
\toprule

\multirow{3}{*}{ \bf Models} & \bf RTE & \bf{MRPC} & \bf{STS-B} & \bf{STS14} & \bf{WikiQA} & \bf{SICK} & \bf{PIT} & \bf{URL} & \bf{TrecQA} & \bf{QQP} & \bf{MNLI} & \bf{SNLI} \\ 
 & 2.5k & 3.5k & 5.7k & 8k & 8k & 10k & 11k & 42k & 53k & 363k & 392k & 549k\\ 
  & Acc & F$_1$ & $r$/$\rho$ & $r$ &MAP/MRR & Acc & max\_$\text{F}_1$ & max\_$\text{F}_1$ & MAP/MRR & Acc & Acc\_m/Acc\_mm & Acc \\ 
 \hline
\(\text{BERT}\) & 65.3 & 88.2 & 86.7/85.8 & 83.6 & 81.8/83.0 & 86.2 & 75.0 & \textbf{78.7} & 84.4/\textbf{89.6} & 90.8 & 84.8/83.1 & \textbf{90.5} \\  
\(\text{BERT}\) + Aligner & \textbf{67.3} & \textbf{88.9} & \textbf{86.8}/\textbf{86.0} & \textbf{83.7} & \textbf{83.2}/\textbf{84.4} & \textbf{87.2} & \textbf{75.5} & 78.5 & \textbf{85.1}/87.8 & \textbf{90.9} & 84.8/\textbf{83.5} & 90.4 \\ 
\bottomrule
\end{tabular}
}
\caption{\label{results_sentence_pair_task} Downstream applications on natural language inference (RTE, SICK, MNLI, SNLI), paraphrase identification (MRPC, PIT, URL, QQP), question answering (WikiQA, TrecQA), and semantic textual similarity (STS-B, STS14) tasks. The datasets in this table are ordered by the size of their training set, as shown in the second row. }
\vspace{-0.5cm}
\end{table*}

\section{Downstream Applications}
\label{sec:downstream}
In this section, we apply our monolingual word aligner to some downstream applications, including both generation and understanding tasks.

\subsection{Automatic Text Simplification}
\label{sec:text_simplification}
Text simplification aims to improve the readability of text by rewriting complex sentences with simpler language. We propose to incorporate word alignment information into the state-of-the-art EditNTS model \cite{dong-etal-2019-editnts} to explicitly learn the edit operations, including addition, deletion and paraphrase. The EditNTS model uses a neural programmer-interpreter architecture, which derives the ADD, KEEP and DELETE operation sequence based on the edit-distance measurements during training time. We instead construct this edit sequence based on the neural semi-CRF aligner's outputs (trained on MTRef$_{Sure+Poss}$) with an additional REPLACE tag to train the EditNTS model (more details in Appendix \ref{sec:editnts_aligner_details}).

Table \ref{results_text_simplification} presents the text simplification results on two benchmark datasets, Newsela-auto and Wikipedia-auto \cite{jiang2020neural}, where we improve the SARI score \cite{xu-etal-2016-optimizing} by 0.9 and 0.6, respectively. The SARI score averages the F$_1$/precision of n-grams inserted (\textbf{add}), kept (\textbf{keep}) and deleted (\textbf{del}) when compared to human references. We also calculate the BLEU score with respect to the input (\textbf{s-BL}), the percentage of new words (\textbf{\%new}) added, and the percentage of system outputs being identical to the input (\textbf{\%eq}) to show the paraphrasing capability. We manually inspect 50 sentences sampled from Newsela-auto test set and find that both models (EditNTS and EditNTS+Aligner) generate the same output for 10 sentences. For the remaining 40 sentences, the original EditNTS only attempts to paraphrase 4 times (2 are good). Our modified model (EditNTS+Aligner) is more aggressive, generating 25 paraphrases (11 are good). With the help of word aligner, the modified model also produces a higher number of good deletions (20 vs. 13) and a lower number of bad deletions (6 vs. 12), which is consistent with the better \textbf{keep} and \textbf{del} scores.

\subsection{Sentence Pair Modeling}
We can utilize our neural aligner in sentence pair classification tasks \cite{lan-xu-2018-neural}, adding conditional alignment probability \(p(\bm{a}|\bm{s, t})\) as an extra feature. We concatenate it with the \texttt{[CLS]} representation in fine-tuned BERT and apply the softmax layer for prediction. We experiment with on different datsets for various tasks, including: natural language inference on SNLI \cite{bowman-etal-2015-large}, MNLI \cite{N18-1101}, SICK \cite{marelli-etal-2014-sick}, and RTE \cite{giampiccolo-etal-2007-third} from the GLUE benchmark \cite{wang2018glue}; semantic textual similarity on STS-B \cite{cer2017semeval} and STS14 \cite{agirre-EtAl:2014:SemEval}; question answering on WikiQA \cite{yang-yih-meek:2015:EMNLP} and TrecQA \cite{wang2007jeopardy}; paraphrase identification on MRPC \cite{dolan2005automatically}, URL \cite{lan2017continuously}, PIT \cite{xu2015semeval}, and QQP \cite{Quora}.

We implement the fine-tuned BERT$_\text{base}$ model using Huggingface's library \cite{Wolf2019HuggingFacesTS}. Table \ref{results_sentence_pair_task} shows performance improvement on small (2k-15k) datasets, which include SICK, STS-B, MRPC, RTE, WikiQA, and PIT, but little or no improvement on large (40k-550k) datasets, such as SNLI, MNLI, and QQP. We hypothesize that the Transformer model can potentially learn the latent word alignment through self-attentions, but not as effectively for small data size.

\section{Conclusion}
In this work, we  present the first neural semi-CRF word alignment model which achieves competitive performance on both in-domain and out-of-domain evaluations. We also create a manually annotated \textbf{Multi}-Genre \textbf{M}onolingual \textbf{W}ord Alignment (MultiMWA) benchmark which is the largest and of higher quality compared to existing datasets. 

\section*{Acknowledgement}
We thank Yang Chen, Sarthak Garg, and anonymous reviewers for their helpful comments. We also thank Sarah Flanagan, Yang Zhong, Panya Bhinder, Kenneth Kannampully for helping with data annotation. This research is supported in part by the NSF awards IIS-2055699, ODNI and IARPA via the BETTER program contract 19051600004, ARO and DARPA via the SocialSim program contract W911NF-17-C-0095, and Criteo Faculty Research Award to Wei Xu. The views and conclusions contained herein are those of the authors and should not be interpreted as necessarily representing the official policies, either expressed or implied, of NSF, ODNI, IARPA, ARO, DARPA or the U.S. Government. The U.S. Government is authorized to reproduce and distribute reprints for governmental purposes notwithstanding any copyright annotation therein.

\bibliography{acl2021}
\bibliographystyle{acl_natbib}

\appendix

\clearpage

\section{EditNTS with Aligner}
\label{sec:editnts_aligner_details}

\begin{table*}[ht!]
  \small
    \centering
    \renewcommand{\arraystretch}{1.1}
    \begin{tabular}{p{0.95\linewidth}}
    \toprule
    Complex sentence:  \\ 
    $[$`With',`Canadian', `collaborators,', `Lloyd', `went', `on', `to', `conduct', `laboratory', `simulations', `of', `his', `model.'$]$ \\
    \hline
    Simple sentence: \\
    $[$`Lloyd', `performed', `successful', `laboratory', `experiments', `of', `his', `model.'$]$ \\
    \hline
    Expert program from EditNTS: \\
    $[$DEL, DEL, DEL, KEEP, ADD(`performed'), ADD(`successful'), DEL, DEL, DEL, DEL, `KEEP', \textcolor{red}{ADD(`experiments'), DEL}, KEEP, KEEP, KEEP$]$ \\
    \hline
    Expert program from EditNTS with Aligner: \\
    $[$DEL, DEL, DEL, KEEP, ADD(`performed'), ADD(`successful'), DEL, DEL, DEL, DEL, `KEEP', \textcolor{red}{`REPLACE-S', ADD(`experiments'), `REPLACE-E'}, KEEP, KEEP, KEEP$]$ \\
    \bottomrule
    \end{tabular}
    \caption{Expert program comparison between the original EditNTS and our modified version with word alignment for the example in Figure \ref{fig:word_alignment_samples}. }
    \label{tab:expert_program_comparison}
\end{table*}

\begin{table*}[!ht]
\small
\centering
\resizebox{.95\textwidth}{!}{%
\renewcommand{\arraystretch}{1.3}
\begin{tabular}{ lcccccccccccc } 
\toprule
\textbf{Models} & \bf SARI & \bf add & \bf keep & \bf del & \bf FK & \bf SLen & \bf OLen & \bf CR & \bf \%split & \bf s-BL &\bf \%new & \bf \%eq \\
\hline
 Complex (input) & 11.8 & 0.0 & 35.5 & 0.0 & 12.3 & 24.8 & 24.8 & 1.0 & 2.0 & 100.0 & 0.0 & 100.0 \\
 Simple (reference) & 86.9 & 84.7 & 78.4 & 97.6 & 6.5 & 13.3 & 13.3 & 0.63 & 0.8 & 25.7 & 33.5 & 0.0 \\
\hline
 EditNTS (original) & 36.6 & 1.1 & 32.9 & 75.7 & 7.5 & 14.3 & 14.3 & 0.66 & 2.4 & 50.2 & 6.5 & 1.2  \\
 EditNTS (original) + Aligner & 36.6 & 1.2  & 32.6 & 75.9 & 7.4 & 14.1 & 14.1 & 0.65 & 2.2 & 49.3 & 6.0 & 1.5  \\
 EditNTS (new) &  36.9 &  1.2 &  \textbf{33.8} &  75.8 &  8.3 & 16.3 & 16.3 & 0.73 & 1.6 & 56.5 & 5.7 & 0.7  \\
 EditNTS (new) + Aligner &  \textbf{37.5} &  \textbf{1.3} &  33.4 &  \textbf{77.9} &  7.2 & 14.3 & 14.3 & 0.66 & 1.5 & 49.1 & 7.6 & 0.8  \\
\bottomrule
\end{tabular}
}
\caption{\label{tab:editnts_model_comparison} Comparison experiments on Newsela-auto dataset with different versions of EditNTS model. + Aligner means using the neural semi-CRF aligner output, EditNTS (new) means adding the REPLACE-S/E tags to the original EditNTS model.}
\end{table*}

The original EditNTS model constructs expert program with the shortest edit path from complex sentence to simple sentence, specifically, it calculates the Levenshtein distances without substitutions and recovers the edit path with three labels: ADD, KEEP and DEL. Since edit distance relies on word identity to match the sentence pair, it cannot produce lexical paraphrases (e.g. \(\textit{conduct} \leftrightarrow \textit{performed}\) and \(\textit{simulations} \leftrightarrow \textit{experiments}\) in Figure \ref{fig:word_alignment_samples},). The final edit sequence will mix paraphrase words (\textit{performed} and \textit{experiments}) and normal added words (\textit{successful}) together under the same ADD label. In order to differentiate these two types of added words, we introduced special tags (REPLACE-S and REPLACE-E) to 
refer to lexical paraphrases specifically. During the edit label construction process, after checking the word pair identity for KEEP label, we additionally check whether they are aligned by our neural semi-CRF aligner, if so, we produce REPLACE-S/E tags, otherwise we do normal ADD/DEL tags. See Table \ref{tab:expert_program_comparison} for a specific example. Word alignment can arbitrarily align any words in the target sentence, this can break the sequential dependency of the edit labels, we therefore discard some lexical paraphrases to guarantee such propriety (\(\textit{conduct} \leftrightarrow \textit{performed}\) in Table \ref{tab:expert_program_comparison}).

In order to show the effectiveness of our modified model, we compared two more versions of EditNTS in Table \ref{tab:editnts_model_comparison}: EditNTS (original) + Aligner, where we directly add word alignment information to the original EditNTS model without any REPLACE tags; EditNTS (new), where we keep the REPLACE tags but don't use any word alignments. The results show that EditNTS model with REPLACE tags can improve the performance, but it is not significant. After adding the word alignment information, we can further improve the SARI score significantly, which can demonstrate the effectiveness of our modified EditNTS with aligner.

\section{More Details for MultiMWA Benchmark}

\subsection{Updated Annotation Guideline}
\label{sec:update_guideline}
After the first round of annotation, we discovery that the definition of phrasal alignment can be ambiguous, which will hinder the development and error analysis for word alignemnt models. Therefore,  we further extend the standard 6-page annotation guideline\footnote{\url{http://www.cs.jhu.edu/~ccb/publications/paraphrase_guidelines.pdf}} from \cite{callison2006annotation} to cover three linguistics phenomena to improve the phrase-level annotation consistency. 

\begin{itemize}
\setlength\itemsep{-0.2em}
    \item ``a/an/the + noun'' should be aligned together with  noun if both nouns are same.
    \item noun$_1$ should be only aligned to noun$_1$ in the phrase ``noun$_1$ and noun$_2$''.
    \item noun should be only aligned to noun in the ``adjective + noun'' phrase.
\end{itemize}

Utilizing the constituency parser implemented in the AllenNLP package \cite{gardner-etal-2018-allennlp}, we first write a script to implement these rules and apply them to all the training/dev/test sets of MultiMWA-MTRef. Then, we manually go through both dev and test sets to further ensure the annotation consistency.

\subsection{Statistics of Alignment Shape}
\label{sec:alignment_shape}
We also analyze the shape of alignment in each dataset, and  the statistics can be found in Table \ref{tab:shape}. Statistical result showes that the dev and test of MultiMWA-MTRef contain a similar portion of phrasal alignment, and less than the training set. There even exists 1$\times$10 alignment annotations in MultiMWA-MTRef, which are actually correct based on our manual inspection. Both MultiMWA-Newsela and MultiMWA-arXiv contain significantly larger portion of 1$\times$1 alignment, especially the latter one contains only 3.2\% of phrasal alignment.

\setlength{\aboverulesep}{0.5pt}
\setlength{\belowrulesep}{0.5pt}

\begin{sidewaystable}
\vspace{-3cm}
\renewcommand\arraystretch{1.25}
\resizebox{\textwidth}{!}{
\begin{tabular}{l|l|c|c|c|ccccccccc|ccccccccc|ccccccc|ccccc|ccc|c} \toprule
                                &       & \textbf{\#pairs} & \textbf{1$\times$1} & \textbf{\%of 1$\times$1} & \textbf{1$\times$2} & \textbf{1$\times$3} & \textbf{1$\times$4} & \textbf{1$\times$5} & \textbf{1$\times$6} & \textbf{1$\times$7} & \textbf{1$\times$8} & \textbf{1$\times$9} & \textbf{1$\times$10} & \textbf{2$\times$2} & \textbf{2$\times$3} & \textbf{2$\times$4} & \textbf{2$\times$5} & \textbf{2$\times$6} & \textbf{2$\times$7} & \textbf{2$\times$8} & \textbf{2$\times$9} & \textbf{2$\times$10} & \textbf{3$\times$3} & \textbf{3$\times$4} & \textbf{3$\times$5} & \textbf{3$\times$6} & \textbf{3$\times$7} & \textbf{3$\times$8} & \textbf{3$\times$9} & \textbf{4$\times$4} & \textbf{4$\times$5} & \textbf{4$\times$6} & \textbf{4$\times$8} & \textbf{4$\times$9} & \textbf{5$\times$5} & \textbf{5$\times$7} & \textbf{5$\times$8} & \textbf{6$\times$6} \\ \hline

\multirow{3}{*}{\textbf{MultiMWA-MTRef}} & Train & 2398    & 30371 & 59.50\%   & 7956 & 3423 & 1648 & 530 & 198 & 112 & 48  & 0   & 10   & 1048 & 1536 & 936 & 270 & 204 & 84  & 16  & 18  & 20   & 549 & 612 & 360 & 198 & 63  & 24  & 27  & 176 & 180 & 72  & 32  & 72  & 25  & 70  & 80  & 72  \\
                                & Dev   & 800     & 10350 & 65.65\%   & 2434 & 876  & 408  & 100 & 30  & 14  & 16  & 9   & 10   & 308  & 474  & 144 & 60  & 48  & 0   & 0   & 0   & 0    & 135 & 156 & 105 & 0   & 0   & 0   & 0   & 48  & 40  & 0   & 0   & 0   & 0   & 0   & 0   & 0   \\
                                & Test  & 800     & 10329 & 63.17\%   & 2650 & 1092 & 404  & 175 & 36  & 28  & 24  & 0   & 0    & 312  & 516  & 200 & 50  & 72  & 0   & 0   & 0   & 0    & 81  & 144 & 75  & 36  & 0   & 0   & 0   & 64  & 40  & 24  & 0   & 0   & 0   & 0   & 0   & 0   \\ \hline
\multirow{3}{*}{\textbf{MultiMWA-Wiki}} & Train & 2514    & 66838 & 95.38\%   & 730 & 315 & 140 & 75 & 36 & 21 & 8  & 0   & 0   & 216 & 258 & 192 & 160 & 36 & 56  & 32  & 0  & 0   & 99 & 168 & 75 & 72 & 42  & 96  & 27  & 32 & 80 & 24  & 56  & 32  & 36  & 0  & 0  & 0  \\
                                & Dev   & 533     & 14499 & 96.13\%   & 200 & 93  & 20  & 15 & 24  & 7  & 0  & 0   & 0   & 32  & 30  & 48 & 40  & 12  & 14   & 0   & 0   & 0    & 36 & 12 & 0 & 0   & 0   & 0   & 0   & 0  & 0  & 0   & 0   & 0   & 0   & 0   & 0   & 0   \\
                                & Test  & 1052     & 28672 & 96.32\%   & 340 & 171 & 72  & 20 & 6  & 14  & 0  & 0   & 0    & 80  & 102  & 32 & 10  & 0  & 0   & 0   & 0   & 0    & 63  & 48 & 0  & 18  & 0   & 0   & 0   & 48  & 0  & 0  & 32   & 0   & 0   & 0   & 40   & 0   \\ \hline
\multicolumn{2}{l|}{\textbf{MultiMWA-Newsela}}    & 500     & 8464  & 74.82\%   & 838  & 483  & 180  & 80  & 48  & 35  & 8   & 36  & 0    & 152  & 276  & 144 & 40  & 24  & 28  & 16  & 18  & 0    & 117 & 180 & 45  & 18  & 63  & 0   & 0   & 0   & 20  & 0   & 0   & 0   & 0   & 0   & 0   & 0   \\ \hline
\multicolumn{2}{l|}{\textbf{MultiMWA-arXiv}}      & 200     & 5020  & 96.80\%   & 78   & 30   & 8    & 0   & 0   & 0   & 0   & 0   & 0    & 8    & 12   & 0   & 0   & 0   & 0   & 0   & 18  & 0    & 0   & 12  & 0   & 0   & 0   & 0   & 0   & 0   & 0   & 0   & 0   & 0   & 0   & 0   & 0   & 0   \\ \toprule
\end{tabular}} 
\caption{Statistics of alignment shapes in each dataset. Each number represents  how many word alignments are included for  phrasal alignment with specific shape. For example, one 2$\times$3 phrasal alignment  will contribute to six word alignments. \%of 1$\times$1 is calculated by 1$\times$1 over the sum of the row. } 
\label{tab:shape}
\end{sidewaystable}

\end{document}